\pdfoutput=1
\documentclass[11pt]{article}

\usepackage{emnlp2021}

\usepackage{times}
\usepackage{latexsym}

\usepackage[T1]{fontenc}
\usepackage[utf8]{inputenc}

\usepackage{microtype}
\usepackage{multirow}
\usepackage{latexsym}
\usepackage{booktabs}
\usepackage{courier}
\usepackage{amsmath}
\usepackage{subfig}
\usepackage{placeins}
\usepackage{tikz}
\usepackage{hyperref}
\usepackage{lipsum}
\usepackage{etoolbox}
\usepackage{comment}

\title{On the Challenges of Evaluating Compositional Explanations in Multi-Hop Inference: Relevance, Completeness, and Expert Ratings}

 \author{Peter A. Jansen \and  Kelly Smith \and Dan Moreno \and Huitzilin Ortiz \\
        University of Arizona, USA \\ \texttt{pajansen@arizona.edu} }
\begin{document}
\maketitle
\begin{abstract}
Building compositional explanations requires models to combine two or more facts that, together, describe why the answer to a question is correct.  Typically, these ``multi-hop'' explanations are evaluated relative to one (or a small number of) gold explanations.  In this work, we show these evaluations substantially underestimate model performance, both in terms of the \textit{relevance} of included facts, as well as the \textit{completeness} of model-generated explanations, because models regularly discover and produce valid explanations that are different than gold explanations.  To address this, we construct a large corpus of 126k domain-expert (science teacher) relevance ratings that augment a corpus of explanations to standardized science exam questions, discovering 80k additional relevant facts not rated as gold.  We build three strong models based on different methodologies (generation, ranking, and schemas), and empirically show that while expert-augmented ratings provide better estimates of explanation quality, both original (gold) and expert-augmented automatic evaluations still substantially underestimate performance by \textit{up to 36\%} when compared with full manual expert judgements, with different models being disproportionately affected. This poses a significant methodological challenge to accurately evaluating explanations produced by compositional reasoning models.
%\todo{\lipsum[1]}
\end{abstract}

\section{Introduction}

%
%   Figure: Motivation
%
\begin{figure}[t]
\centering
\includegraphics[width=1.00\columnwidth]{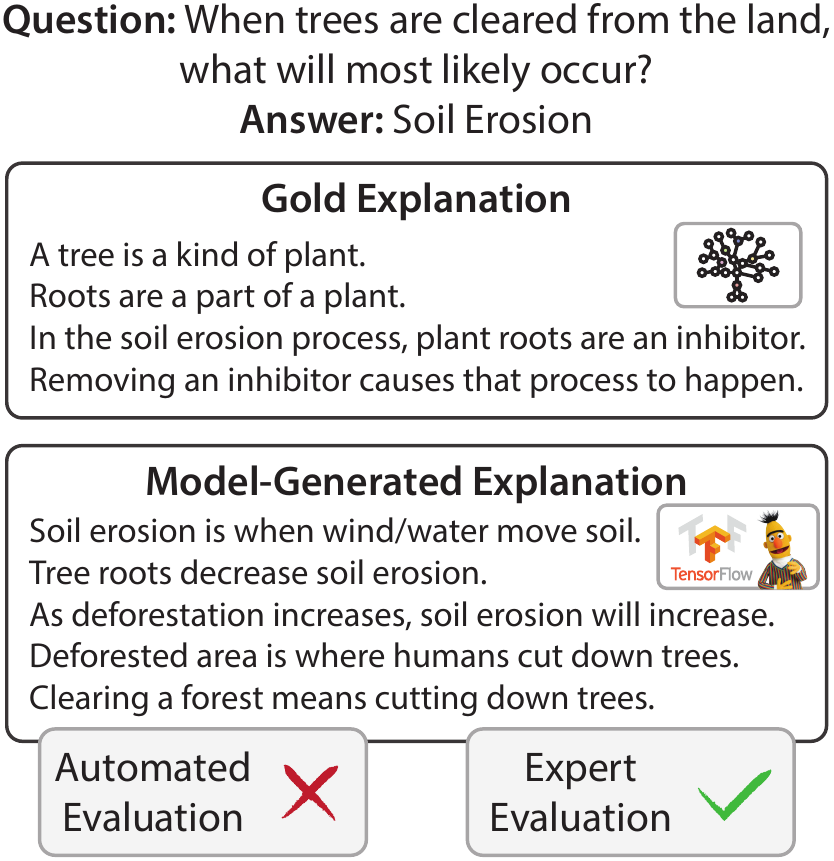}
\caption{\footnotesize An example science exam question, its gold explanation from the WorldTree corpus, and a model-generated explanation from one of the models \textit{(Tensorflow-Ranking-BERT)} trained using expert-generated relevance ratings produced in this work.  Though the model-generated explanation is strong, it shares no facts in common with the gold explanation, and automatic evaluations rate it neither \textit{relevant} nor \textit{complete}. }%\todo{(PJ note: possible to show relevance ratings on here? old vs new?)}}
\label{fig:motivation}
\vspace{-4mm}
\end{figure}

% Super high-level introduction to the research area
Compositional inference is the high-level task of combining two or more pieces of knowledge to perform reasoning. % elaborate
In the context of question answering, compositional (or ``multi-hop'') inference typically takes the form of combining facts from a knowledge base that allow a given solver to form a complete chain-of-reasoning that moves from question to correct answer.  A desirable consequence is that the facts used to assemble this chain-of-reasoning can then be taken as an interpretable record of that reasoning, as well as a human-readable explanation for why the answer is correct. 

Compositional inference has seen steady growth in the last three years, in large part due to the recent availability of training and evaluation data for the task \citep[e.g.][]{yang-etal-2018-hotpotqa,khashabi-etal-2018-looking,jansen-etal-2018-worldtree}, which has historically been unavailable due to the challenges in annotating explanations, and the expense in generating quality data at scale.  To ease these burdens, nearly all datasets have focused on small compositional inference problems that require composing only two representations, typically triples, sentences, or whole paragraphs \cite[see][for a survey of datasets]{wiegreffe2021teach}.

In this work, we focus on the problem of generating and evaluating large explanations to science exam questions (with the average explanation in this work requiring composing 6 facts).  Our evaluation experience in this domain has been uneasy -- we have developed seemingly well-reasoned models, only to receive comparatively low evaluation scores relative to baseline models in automatic evaluations.  When evaluated manually, as shown in Figure~\ref{fig:motivation}, we observe that many models produce compelling explanations, at least in part, but these explanations score poorly because they differ from gold explanations. This parallels the disparity in automatic versus manual evaluations in other fields, such as machine translation \cite{freitag2021experts}.

In this work we systematically analyze the difference between automatic and expert manual evaluation, and formalize evaluation of large compositional explanations in two aspects: by examining the \textbf{relevance} of each fact to the question and answer, as well as the \textbf{completeness} of the entire explanation -- that is, whether the collection of facts in the explanation form a complete chain-of-reasoning from question to answer.  % Formalize this a bit more?
Of these two metrics, obtaining accurate \textit{relevance} measures is in principle solvable by brute force -- by creating an exhaustive corpus of ratings -- while exhaustively enumerating possible \textit{n-fact} explanations and rating them for \textit{completeness} is likely less tractable. 
%for each question, rate the relevance a given fact from a knowledge base might have when used in an explanation for the answer -- and in this work we focus on generating and analyzing an approximate of this form of exhaustive evaluation.  Conversely, evaluating \textit{completeness} exhaustively is more challenging, and in principal expands following the binominal coefficient as all combinations of \textit{n} facts are chosen from a corpus then evaluated for completeness.  
Here, we focus on generating extensive \textit{relevance} annotation of the facts most likely to be incorporated in explanations, while providing an estimate of the undercounting of \textit{completeness} by manually evaluating the completeness of explanations generated by three state-of-the-art models. 

% made by domain-experts (science teachers).
The contributions of this work are: 
\begin{enumerate}
    \item \textbf{Resource:} We produce a large set of 126k expert-generated relevance ratings for building explanations to science exam questions from atomic facts. Each rated fact was highly ranked by a large language model trained on gold explanations using distant supervision, complementing those in the WorldTree V2 explanation corpus \cite{xie-etal-2020-worldtree}.  The domain experts (science teachers) discovered 80.6k additional facts, \textit{four times more than provided in gold explanations}, to be relevant for explanation construction. %. , unknown to gold explanations but relevant for explanation construction, nearly four times those originally annotated. 
    \item \textbf{Models:} We use these ratings to train and evaluate three state-of-the-art exhaustive models, using three different modeling paradigms: explanation-as-ranking, generation, and constraint-based schemas.
    \item \textbf{Evaluation:} We conduct large automatic and manual evaluations, empirically demonstrating substantial differences in evaluation when using better ratings and judgements. In fully-automatic evaluations, original evaluations underestimate \textit{relevance} by up to 14\% compared with a fully-automatic evaluation that includes expert ratings.  But, this fully-automatic setting still underestimates \textit{relevance} by up to 29\% and \textit{completeness} by up to 36\% compared to full manual judgements. 
\end{enumerate}

%
%   Figure: Overview
%
\begin{figure*}
\centering
\includegraphics[width=1.95\columnwidth]{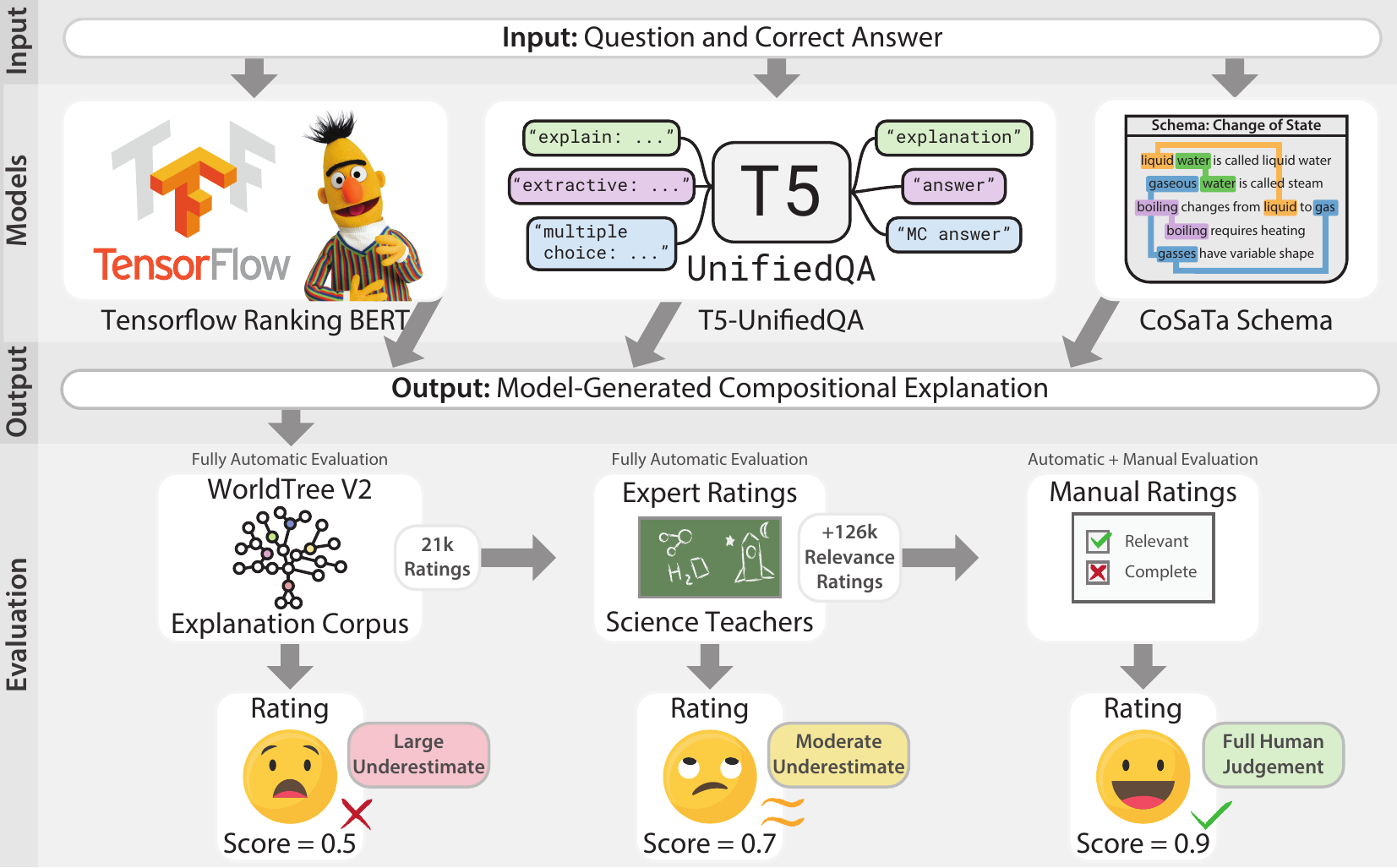}
\caption{\footnotesize An overview of this work.  We generate a large set of relevance ratings for explanatory facts annotated by domain experts (science teachers), that complement the original WorldTree explanation corpus.  We use this new annotation for training and evaluating three families of strong models (Tensorflow Ranking BERT, T5-UnifiedQA, and CoSaTa Schemas) on generating explanations.  We show through automatic and manual analyses that the current method of using a single gold explanation for evaluation substantially undercounts explanation performance in terms of relevance and completeness, while even expert relevance ratings (when used in fully-automatic evaluations) still moderately undercount true task performance compared to full manual human judgements. }
\label{fig:overview}
%\vspace{-4mm}
\end{figure*}

% Contributions
%\todo{\lipsum{1]}

%\subsection{Contribution}
%\begin{itemize}
%    \item Release of a new resource for the evaluation of explanatory relevance in the context of multi-hop explanation generation for science questions
%    \item Data creation and collection methodology
%    \item Exhaustive evaluation of different language models with focus on their capability to capture explanatory relevance
%    \item Empirically show that the new data, when integrated in Worldtree, can be used to fine-tune and improve language models on the relevance task while maintaining good performance in explanation regeneration.
%    \item Show the impact of the new data on downstream question answering. The hypothesis is that explanation retrieval models fine-tuned on the explanatory relevance dataset can provide better evidence for downstream QA models. 
%\end{itemize}

\section{Related Work}

%%
%%   Figure: Spectrum
%%
%\begin{figure}
%\centering
%\includegraphics[width=\columnwidth]{figures/spectrum-3.pdf}
%\caption{placeholder}
%\label{fig:spectrum}
%\end{figure}

%\todo{ \lipsum[1-4] }

\paragraph{Compositional explanations:} In their survey, Weigreffe and Marasovic~\shortcite{wiegreffe2021teach} identified 14 structured explanation datasets for compositional reasoning.  Due to the challenge in annotating large compositional explanations, nearly all datasets to date (such as QASC \cite{khot2020qasc}, OpenBookQA \cite{mihaylov2018can}, and R$^4$C \cite{inoue-etal-2020-r4c}) require combining an average of only 2 facts.  In this work, to study evaluation challenges with the largest available explanations, we use the WorldTree V2 explanation corpus \cite{xie-etal-2020-worldtree}, whose explanations require composing an average of 6 (and as many as 16) facts.

\paragraph{Evaluation with multiple gold explanations:} Nearly all compositional reasoning datasets annotate (at most) a single gold explanation, with two exceptions.  eQASC \cite{jhamtani-clark-2020-learning} generates 10 perturbations of possible 2-fact QASC explanations, and asks crowdworkers to rate these as valid or invalid chains of reasoning.  26\% are rated valid, resulting in an average of 2.6 valid 2-fact explanations per question, which Jhamtani and Clark~\shortcite{jhamtani-clark-2020-learning} then use to train a classifier and ranker.  Taking a different approach, R$^4$C \cite{inoue-etal-2020-r4c} uses crowdworkers to generate 3 explanations (represented as chains of triples) to select HotpotQA questions \cite{yang-etal-2018-hotpotqa}. R$^4$C explanations are short, with 68\% containing 2 triples, 23\% using 3 triples, and 9\% use 4 or more triples. Inoue et al.~\shortcite{inoue-etal-2020-r4c} then define an alignment procedure between model-generated output triples and the 3 gold explanations, and take the highest scoring alignment as the score of the explanation.  In this work, due to the intractability of generating and rating explanatory perturbations with large explanations, we instead use domain-experts to produce relevance ratings for component facts at scale.  %We also show a large gap in automatic vs manual completeness ratings due to the large number of possible valid large explanations, suggesting that 
We then rate model explanations for \textit{completeness} manually, as we hypothesize that generating a large database of alternative gold explanations as in Inoue et al.~\shortcite{inoue-etal-2020-r4c} is likely intractable for explanations longer than two or three facts. 

%\paragraph{Posthoc explanations:} \todo{(Omit for space?)} Explanations for an inference can be generated as a faithful record of the actual reasoning process used to arrive at an answer (e.g. \todo{(cite)}), or as a post-hoc rationalization for why a given answer is correct.  Given that the former requires explainable models that are amenable to introspection, and most contemporary approaches to explanation regeneration do not have this property (e.g. \todo{(cite)}), in this work we focus on post-hoc explanations as a stepping stone. 

%% Cut for space
%\paragraph{Relevance and completeness:} In a small study, Jansen and Ustalov \cite{jansen-ustalov-2019-textgraphs} used manual measurements on 14 questions to estimate that relevance may be significantly undercounted in explanation-as-ranking paradigms.  Here, we systematically study both \textit{relevance} and \textit{completeness} in  explanation-as-ranking and whole-explanation paradigms, using two orders of magnitude more data, generated by domain experts.  %To the best of our knowledge, this is the first study of completeness in large model-generated compositional explanations, but our ratings are informed by work understanding explanatory completeness in other domains (e.g. psychology). 

Rating completeness is challenging, with opportunities for bias. For example, Korman et al.~\shortcite{korman2020explanatory} note that ``all explanations are incomplete, but reasoners think some explanations are more complete than others'', and empirically determined that humans prefer simpler explanations that reduce gaps in causal explanatory steps.  Here we consider explanations correct if they are technically correct (as determined by domain experts), and minimize gaps in inferences, without measuring succinctness. 

\paragraph{Modeling approaches to generating explanations:} A wide variety of approaches have been proposed for building compositional explanations \citep[see][inter alia]{thayaparan2020survey}, including integer linear programming \cite{khashabi2016question}, formal logics and rules \cite{weber-etal-2019-nlprolog}, iterative construction methods \cite{cartuyvels-etal-2020-autoregressive}, and various explanation-as-ranking approaches such as those proposed in the Shared Tasks for Explanation Regeneration \citep[e.g.][]{jansen-ustalov-2020-textgraphs}.  In this work we explore evaluation challenges grounded in the performance of three strong and methodologically diverse models: reranking exhaustive classifications of large language models \citep[e.g.][]{das2019chains,li2020pgl}, generative models \citep[e.g.][]{khashabi-etal-2020-unifiedqa}, and schema-based models \citep[e.g.][]{lin-etal-2019-kagnet,jansen-2020-cosata}

%Nearly all top-scoring approaches to generating large compositional explanations combine a strong retrieval component with reranking or other iterative methods to distill a candidate list of facts into a final explanation (see \cite{thayaparan2020survey} for a review). While large language models show strong performance across a variety of models in learning-to-rank paradigms \cite{pawate2020chisquarex}, Das et al. ~\shortcite{das2019chains} show that exhaustively training and evaluating BERT as a fact-level classifier for every fact in a knowledge base, while computationally expensive, can yield impressive performance alone. This has since been extended, and Li et al.~\shortcite{li2020pgl} have shown incremental performance as the number of parameters in the model increases.  \todo{Discuss generative/schema models?} In this work, we explore the performance of strong exhaustive models from three modeling paradigms: ranking models, generative models, and schema-based models. 

% cite EMNLP 2019 paper on using schemas (that were just triples?)

%\todo{XX} et al. demonstrate that a QA model that selects multiple passages by maximizing term frequency as a measure of relevance, and inverse document frequency as a measure of diversity, can perform well at compositional retrieval. 

%\section{Approach}
%-+
%\todo{TODO: Figure for approach.}
%
%\subsection{Questions}
%WorldTree, ARC, etc. 
%
%Set sizes. 
%
%\subsection{Evaluation Metrics}
%
%- Ranking: NDCG, P@K
%
%- Classification: ..., new metric (P+R~=F1)
%
%- Generation: Align using ROUGE, + Manual Analysis
%
%- Manual Analyses.

\section{Overview}
An overview of our approach is shown in Figure~\ref{fig:overview}.  First, in Section~\ref{sec:dataset} we generate a large set of expert relevance ratings that complement those in the WorldTree V2 explanation corpus.  We then use these expert relevance ratings to demonstrate existing evaluations substantially undercount relevance in explanation-as-ranking paradigms in Section~\ref{sec:explanation-as-ranking}.  In Section~\ref{sec:experiments} we implement strong generative, ranking, and schema-based models that produce whole explanations (rather than ranked lists), and show through manual analysis that automated metrics substantially undercount relevance and completeness of whole explanations, even when using better ratings.  We conclude with a discussion of implications for evaluation in multi-hop inference.

\section{Dataset Description}
\label{sec:dataset}

To support our experimentation we created a resource that, for a given WorldTree V2 question, provides a set of relevance ratings for the facts most likely to be in an explanation.  WorldTree V2 contains 4.4k questions, and its supporting knowledge base contains approximately 9k facts, meaning that exhaustively evaluating the relevance of each fact for each question would require approximately \textit{40 million} ratings, which is intractable even for this comparatively small corpus.  Instead, here, annotators rate a shortlist of facts most likely to be relevant to building possible explanations, producing ratings for a total of 126k facts.

\begin{table}[t!]
\begin{center}
\footnotesize
\setlength{\tabcolsep}{3pt}
\begin{tabular}{ccl} 
\toprule
\multicolumn{3}{c}{\textbf{Q:} Burning fossil fuels adds pollutants like sulphur into } \\
\multicolumn{3}{c}{the air. This pollution contributes to:} \\
\multicolumn{3}{c}{\textbf{A:} acid rain} \\
\midrule
TR  & Gold  &  \multicolumn{1}{c}{Fact} \\
3   &   *   & Burning fossil fuels releases sulfur dioxide \\%/chemical energy) \\% into
    &       &   ~~into the atmosphere.      \\
3   &   *   & Emitting sulfur dioxide causes acid rain. \\
2   &      & Burning fossil fuels causes pollution.  \\
2   &   *    & Emission is when something is added to. \\
    &       & ~~the atmosphere.  \\
2   &       & Gasses from burning oil and coal that dissolve \\
    &       & ~~in water in the atmosphere cause acid rain.\\
2   &       & As the amount of sulphur gas in the atmosphere \\
    &       & ~~increases, the PH of rain will decrease. \\
1   &       & Acid rain negatively impacts water quality. \\
1   &       & Coal is a kind of fossil fuel.  \\
0   &       & The air contains carbon dioxide.  \\
0   &       & Oil is a kind of pollutant. \\
\bottomrule
\end{tabular}
\caption{\footnotesize Example relevance ratings. \textit{(top)} A question and its correct answer. \textit{(bottom)} A subset of the shortlist of facts, teacher-generated relevance ratings (TR) for each fact, and whether a given fact was included in the gold explanation in the original WorldTree V2 explanations. % The relevance rating reflects the sum of two ratings by two independent annotators. }
\textit{Note:} for space, only a subset of the shortlist of facts and gold explanation are shown.
\label{tab:example-relevance-ratings}}
\end{center}
\end{table}

\paragraph{Initial Shortlist:} To produce the shortlist, we train two large language models, BERT \cite{devlin-etal-2019-bert} and RoBERTa \cite{liu2019roberta}, on the task of retrieving relevant sentences from the corpus.  To encourage the models to find a broad array of facts that might be relevant to building an explanation, we model this retrieval as a distant supervision classification problem where the gold WorldTree explanations are used as positive examples, while 200 randomly sampled facts from the corpus serve as negative training examples. %%\footnote{Pilot experiments showed that using all non-gold facts in the corpus as negative training examples produced less favorable performance, as other relevant facts not rated as gold by the original WT21 annotators are then treated as incorrect.}  
For each question, we exhaustively score all 9k facts in the knowledge base using both models, take the top 20 scoring facts from each model, and combine them into a final shortlist.  We also add any facts from the gold explanation that were not ranked in the top 20 by either model. Due to significant overlap in the output of both models, the average shortlist per question contains 28.9 facts.\footnote{Scoring only the top $\approx$30 facts per question was chosen due to budgetary constraints and timing considerations.} %- Also discuss that synonymy facts are removed/not rated?

\paragraph{Rating protocol:} Each question's shortlist of facts was independently rated by 2 domain-expert annotators (science teachers), using the following 4-point rating scheme:

%\begin{itemize}
%    \item \textbf{Core (3):} Facts that directly address the core topic the question is testing, or directly answer the question.
%    \item \textbf{Important (2):} Key knowledge supporting the core facts or grounding core knowledge in examples the question uses.
%    \item \textbf{Extra Detail (1):} Facts that (a) when included, add extra detail to the explanation, but (b) when missing, do not exclude important details from the explanation.
%    \item \textbf{Irrelevant (0):} Facts not relevant to the question.
%\end{itemize}

\begin{table}[h]
\begin{center}
\footnotesize
\setlength{\tabcolsep}{3pt}
\begin{tabular}{clp{4.75cm}} 
\toprule
TR   & Label        & Description   \\
\midrule
3   & Core          & Facts that directly address the core topic the question is testing. \\[1.5mm]
2   & Important     & Key knowledge supporting the core facts or grounding core knowledge in examples the question uses. \\[1.5mm]
1   & Extra Detail  & Facts that (a) when included, add extra detail to the explanation, but (b) when missing, do not exclude important details from the explanation. \\[1.5mm]
0   & Irrelevant    & Facts not relevant to the question. \\
\bottomrule
\end{tabular}
\caption{\footnotesize 4-point Relevance Rating Scheme}
\label{tab:rating-scheme}
\end{center}
\end{table}

Central to this rating scheme is a graded notion of relevance, that includes optional facts \textit{(extra detail)} that enrich the explanation when included, but do not cause critical gaps in the inference when not included.  %, who trained together for approximately \todo{20} hours to establish ratings criteria, then worked independently for several months to complete the ratings for the dataset.  
Example ratings are shown in Table~\ref{tab:example-relevance-ratings}.

\paragraph{Raters:} Three graduate research assistants in education served as domain experts, and worked for several months to complete the relevance ratings.  Each has between 8 and 20 years of science teaching experience at the elementary, middle-school, or high-school level.

\paragraph{Interannotator Agreement:} Even after substantial training, the domain experts found this to be a challenging task. Interannotator agreement (Cohen's Kappa) was $\kappa = 0.46$, which is considered moderate agreement \cite{landis1977measurement}. Raw percent agreement between annotators was 61\%, with nearly all disagreements within $\pm1$ of each other (88\% of disagreements). Annotators reported that disagreements tended to be in determining thresholds for the different categories -- some annotators tended to err on the side of suggesting more facts were important to generating an explanation, while others preferred generating more minimalistic explanations. %\footnote{Annotators also reported that it was challenging to rate facts in isolation of each other, and that they often had to consider different possible collections of facts that might form complete explanations before rating facts individually.  \todo{We discuss this more in Section~\ref{}, Discussion.}}  
To account for individual variation in detail preference, we average the final ratings and round up the final scores.

% Raw data 
%                          GOLD_RATED_1.0	315.0
%                          GOLD_RATED_2.0	2846.0
%                          GOLD_RATED_3.0	9063.0
%                          GOLD_RATED_4.0	8579.0
%                       NOTGOLD_RATED_1.0	24795.0
%                       NOTGOLD_RATED_2.0	36962.0
%                       NOTGOLD_RATED_3.0	36399.0
%                       NOTGOLD_RATED_4.0	7245.0
%                           NUM_QUESTIONS	4374.0
%                          NUM_ROWS_RATED	126204.0
\begin{table}[t!]
\begin{center}
\footnotesize
\begin{tabular}{cccccc} 
\toprule
 ~ & ~       & \multicolumn{4}{c}{Teacher Rating} \\
 ~ & ~       &   0   &   1   &   2   &   3   \\
 ~ & ~       &   Irr. &   Ext.   &   Imp.   &   Core   \\
 \midrule
 \parbox[t]{2mm}{\multirow{2}{*}{\rotatebox[origin=c]{90}{WT2}}} & Gold       & 315   & 2,846  &   9,063    &  8,579  \\ 
 ~ & Not Gold   & 24,795 & 36,962 &   36,399   &  7,245 \\ 
 \midrule
 ~ & Increase   &   --  &   1299\%   &   402\%   &   84\%   \\
\bottomrule
\end{tabular}
\caption{\footnotesize Distribution of teacher ratings for shortlisted facts across all explanations, broken down by each fact's original WorldTree V2 (WT2) rating.  In total, this annotation procedure discovered 80.6k additional relevant facts in the corpus (approximately 18 facts per question) not originally included in the single gold WorldTree explanation per question.
\label{tab:dataset-gold-vs-notgold}}
\end{center}
\end{table}

%
%   Main explanation-as-ranking results table
%
\begin{table*}[t!]
    \footnotesize
    \centering
    \begin{tabular}{lc|c|ccccc|c}
    \toprule
    ~ &                &   \multicolumn{6}{c}{\textbf{Explanation-as-Ranking Evaluation Method (MAP)}}  \\[1mm]
    ~ &  \textbf{Ranking}              &   \textbf{Baseline}       &   \textbf{Teacher}  &           &   \textbf{Teacher}  &        & ~ &  \textbf{Teacher}   \\
    \textbf{Model}  &  \textbf{Setting} & \textbf{WT2}       &   \textit{(>= Extra (1))}  &     $\Delta$   &   \textit{(>=Important (2) )}  &     $\Delta$   &  ~ & \textbf{NDCG}   \\
    \midrule
    % TFIDF & Exhaustive          &   0.31  &   0.20  &   $\downarrow$0.11  &   0.22  &   $\downarrow$0.09  & ~ &   0.50    \\
    BERT & Exhaustive            &   0.54  &   0.66  &   $\uparrow$0.11  &   0.58  &   $\uparrow$0.03  & ~ &   0.75  \\
    RoBERTa & Exhaustive        &   \textbf{0.57}  &   \textbf{0.68}  &   $\uparrow$0.11  &   0.61  &   $\uparrow$0.04  & ~ &   0.78  \\
    TFR-BERT & Rerank (K=100)        &   0.49  &   0.58  &   $\uparrow$0.09  &   \textbf{0.63}  &   $\uparrow$0.14  & ~ &   \textbf{0.81}  \\
    \bottomrule
    \end{tabular}
    \caption{\footnotesize Using more extensive expert relevance ratings causes substantially different MAP scores for the same models in fully-automatic evaluation settings. \textit{(Left)} Models, and their ranking setting.  All models were exhaustively evaluated on the entire corpus except TFR-BERT, which reranks the top-100 BERT facts per question. \textit{(Right)} Evaluation scores when using original WorldTree (WT2) ratings, or the top-K expert (science teacher) ratings produced in this work.  Teacher evaluations are provided in two settings: considering all facts rated ``Extra'' or above as gold, or all facts rated ``Important'' or above as gold.  All scores represent Mean Average Precision (MAP), except for the last column, which provides Normalized Discounted Cumulative Gain (NDCG) for reference.  Delta scores ($\Delta$) represent the difference between a given teacher evaluation and the WT2 baseline.  Automatic measures of relevance can differ by up to 14\% when using either a single gold explanation or top-K expert relevance ratings as gold.}
    \label{tab:ranking-results}
\end{table*}

\paragraph{Comparison with gold explanations:} The distribution of relevance ratings is shown in Table~\ref{tab:dataset-gold-vs-notgold}, broken down by whether a given fact was originally included in the gold WorldTree explanation for a given question.  Of the 20.8k facts across all gold WorldTree explanations, the teachers rated 98.5\% of these as also relevant (i.e., $TR>0$), demonstrating strong agreement with the original explanation authors.  Teachers rated 17.6k (85\%) of these facts as \textit{core} or \textit{important} to the inference, while 2,846 (14\%) were rated as \textit{extra detail} facts that the explanations could include or disclude without causing significant gaps in the reasoning.

Most stark is the volume of additional facts rated as relevant by the domain experts.  In total, 80.6k additional facts (approximately 18 facts per question) were rated as relevant by teachers, but were not used in a given question's gold explanation, with 43.6k of these facts (10 per question) rated as \textit{core} or \textit{important}. As context, the original WorldTree explanations contain an average of 6 facts -- here, the expert ratings found that, on average, four times as many facts are relevant to building an explanation than are annotated in a given question's single explanation.  This suggests that \textit{while providing a single example explanation is helpful for training a model, it is insufficient for evaluating that model's capacity for constructing explanations}, as it may be possible to build many different compositional explanations from a given collection of facts.

%- Rating protocol, 2 annotators per fact
%- Expert annotators (science teachers)
%- Rating shortlist generation: Top 20 ratings from BERT, ROBERTA, trained using distant supervision -- combined into a single list (top ~30). 
%- Total number of ratings
%- Interannotator agreement (percent agreement? kappa? confusion matrix?)
%- Sources of disagreement
%- Agreement with original WT annotation (highlighting how many other possible solutions there may be, given the large number of other possible 'good' ratings).

\section{Experiments: Explanation-as-Ranking}
\label{sec:explanation-as-ranking}

To characterize differences in automatic evaluation when using existing gold explanations versus expert relevance ratings, we first explore performance in the explanation-as-ranking paradigm. Rather than directly producing an explanation for a given question, as a stepping-stone task, explanation-as-ranking \cite{jansen-ustalov-2019-textgraphs,das2019chains,li2020pgl} is an explanatory retrieval analogue that requires models to exhaustively rank all facts in a knowledge base such that the most relevant facts are selectively ranked to the top of the list.\footnote{Note that the dataset generated in this paper was used for the Third Shared Task on Multi-Hop Inference for Explanation Regeneration, an explanation-as-ranking task \cite{jansen-etal-2021-textgraphs} that ran concurrently with this submission.  The TFR-BERT model described here performs comparably with the winning system \cite{pan2021deepblueai}, which reached 0.82 NDCG.}%  As a direct comparison to previous work, we use this as a means of empirically evaluating the difference in evaluation when using original WorldTree scores, or the expert-generated teacher ratings produced in this work. 

\paragraph{Models:} We include the exhaustive BERT and RoBERTa models trained on the original WorldTree gold explanations, and used in generating the shortlist for the teacher ratings, as described above.  We also include a Tensorflow-Ranking-BERT model \cite{han2020learning}, which combines BERT embeddings directly in a pointwise learning-to-rank \cite{pasumarthi2019tf} instead of classification framework, and achieves extremely strong single-model performance for large benchmark ranking tasks such as MS MARCO \cite{nguyen2016ms}. Unlike the baseline models, TFR-BERT was trained on the expert-generated relevance ratings.  Due to the expense in evaluating this model, here we rerank only the top 100 scoring facts from the exhaustive BERT model.  % TODO: Hyperparameters in appendix?

\paragraph{Evaluation:} Explanation-as-ranking performance is reported using Mean Average Precision (MAP).  We evaluate in three settings: (1) Using the original gold WorldTree explanations, (2) treating each fact the experts rated as \textit{extra-detail} or higher as gold, or (3) treating each fact the experts rated as \textit{important} or higher as gold.  Because the expert ratings are graded (0-3) rather than binary (gold/not gold), we also evaluate using Normalized Discounted Cumulative Gain (NDCG). 

\paragraph{Results:} The results of the evaluation are shown in Table~\ref{tab:ranking-results}.  Using the original gold annotation, the best scoring model, RoBERTa, achieves a MAP of 0.57.  When evaluated using the expert relevance ratings, this increases to 0.61 (+4\%) when considering only \textit{important} or higher facts as gold, while increasing to 0.68 (+11\%) when allowing \textit{extra detail} facts to be considered gold.  Conversely, the TFR-BERT shows comparatively low performance using the original gold annotation, at 0.49 MAP.  When evaluated using the exhaustive teacher ratings, this model achieves a MAP of 0.63 (+14\%) when considering only \textit{important} or higher facts as gold -- becoming the best-performing model -- while reaching 0.58 (+9\%) when considering \textit{extra detail} or better facts as gold.  These results empirically demonstrate that using a single gold explanation as an evaluation standard can dramatically underestimate model performance compared to more extensive relevance annotation. Further, \textit{this performance decrease may not be uniform across models}, as demonstrated by RoBERTa's performance being underestimated by only 4\% while TFR-BERT is underestimated by 14\% -- suggesting that meaningful comparisons between model performance may not be possible with limited relevance annotation.

%
%   Main constraints results table (TEST)
%
\begin{table*}[t!]
    \footnotesize
    \centering
    %\ra{1}
    \begin{tabular}{lccccccc}
    \toprule
    & \multicolumn{5}{c}{Automatic Analysis} & Average & \\
    Model & Rel & Comp & ${F1}^{ex}$ & Comp$_{B}$ &  ${F1}^{ex}_{B}$ & Expl. Length \\
    \midrule
    \multicolumn{4}{l}{\textbf{Single Models}} \\
    \midrule
    \multicolumn{4}{l}{\textit{Generative and Ranking}}\\
    ~~T5-UQA-3B$_{CORE}$                  & 0.62    &   0.32    &   0.42    &   0.20 &	0.12 &   8   \\
    ~~T5-UQA-3B$_{CORE+EXT}$              & 0.55    &   0.39    &   0.45    &   0.23 &	0.14 & 13  \\
    ~~TFR-BERT                            & 0.74    &   \textbf{0.59}    &   \textbf{0.66}    &   \textbf{0.50} &	\textbf{0.37} & 8   \\
    ~\\
    %\multicolumn{4}{l}{\textit{Schema-based (unfiltered)}} \\    
    %~~Schema$_{U}$ (1 Schema)           & 0.43  &   0.38    &   0.40    & 0.25	& 0.18 &    12  \\
    %~~Schema$_{U}$ (2 Schemas)          & 0.33  &   0.46    &   0.38    & 0.27	& 0.23 &   23  \\
    %~~Schema$_{U}$ (3 Schemas)          & 0.27  &   0.50    &   0.35    & 0.27	& 0.27 &  31  \\
    %~\\
    %\multicolumn{4}{l}{\textit{Schema-based (filtered)}} \\
    \multicolumn{4}{l}{\textit{Schema-based}} \\    
    ~~Schema (1 Schema)           & \textbf{0.82}   &   0.36    &   0.50    & 0.27	& 0.16 &   5   \\
    ~~Schema (2 Schemas)          & 0.78   &   0.42    &   0.55    & 0.32	& 0.20 &  7   \\
    ~~Schema (3 Schemas)          & 0.75   &   0.46    &   0.57    & 0.36	& 0.23 &   8   \\
    \midrule
    \multicolumn{4}{l}{\textbf{Ensembles}}   \\
    \midrule
    ~~T5-UQA-3B$_{C+E}$ + TFR-BERT      & 0.62  &   0.70    &   0.66    & 0.54	& 0.48 &   21  \\
    %~\\
    %~~Schema$_{F}$ (1S) + TFR-BERT      & 0.77  &   0.67    &   0.72    & 0.56	& 0.45 &   13  \\
    %~~Schema$_{F}$ (2S) + TFR-BERT      & 0.76  &   0.70    &   0.73    & 0.59	& 0.48 &   15  \\
    ~~Schema (3S) + TFR-BERT      & \textbf{0.75}  &   0.71    &   \textbf{0.73}    & 0.59	& 0.49 &   16  \\
    %~\\
    %~~Schema$_{F}$ (1S) + T5-UQA-3B$_{C+E}$   & 0.61    &   0.55    &   0.58    & 0.40	& 0.30 &   19  \\
    %~~Schema$_{F}$ (2S) + T5-UQA-3B$_{C+E}$   & 0.62    &   0.59    &   0.61    & 0.44	& 0.34 &   20  \\
    ~~Schema (3S) + T5-UQA-3B$_{C+E}$   & 0.62    &   0.61    &   0.62    & 0.46	& 0.36 &   22  \\
    %~\\
    %~~Schema$_{F}$ (1S) + TFR-BERT + T5-UQA-3B$_{C+E}$   & 0.65 &   0.74    &   0.70    & 0.59	& 0.53 &   27  \\
    %~~Schema$_{F}$ (2S) + TFR-BERT + T5-UQA-3B$_{C+E}$   & 0.65 &   0.76    &   0.70    & 0.60	& 0.56 &   28  \\
    ~~Schema (3S) + TFR-BERT + T5-UQA-3B$_{C+E}$   & 0.65 &   \textbf{0.77}    &   0.71    & \textbf{0.61}	& \textbf{0.57} &   30  \\
    
    \bottomrule
    \end{tabular}
    \caption{\footnotesize Performance of all single and ensemble models investigated, using automatic performance metrics.  \textit{Relevance} is measured using expert (teacher) ratings, while \textit{completeness} is measured using a combination of orginal WorldTree V2 and teacher ratings (see text).}
    \label{tab:model-results}
\end{table*}

\section{Experiments: Whole Explanations}
\label{sec:experiments}

Here we construct and evaluate a diverse set of generative, top-k ranking, and schema-based models that return whole (short-length) explanations to a user, in place of ranked lists.  We evaluate these in terms of both \textit{relevance} and \textit{completeness}, comparing both automatic and manual assessments of these metrics. 

%\todo{Move to approach?} In this work we compose explanations through generation, top-K ranking, or schema slot filling, which output a collection of facts each model believes to be in an explanation.  As a direct comparison to previous work, e.g. the Shared Tasks on Explanation Regeneration \todo{(cite)}, we also briefly discuss explanation as solely as a ranking problem below.  

\subsection{Models}
Each model is described briefly below, with details and hyperparameters provided in the \texttt{Appendix}. 
%\todo{- All models generate post-hoc explanations, and are cued with the question and correct answer text, which are then used for generating explanations directly (as in T5), scoring and ranking component facts (TFR-BERT), or scoring schemas and their component facts (CoSaTa). }

\paragraph{T5-UnifiedQA (Generative):} UnifiedQA \cite{khashabi-etal-2020-unifiedqa} is a variant of T5 \cite{2020t5} that includes multi-task pretraining for 4 different forms of question answering across 20 datasets, including extractive QA (i.e., locating an answer span in a passage), abstractive QA (i.e., generating an answer not supplied in a passage), multiple-choice QA, and Boolean QA, while achieving state-of-the-art performance on 10 datasets.  Here we train T5-UQA-3B to also generate compositional explanations by cueing generation with the question and correct answer candidate, and targeting generation to produce strings of highly-rated facts delimited with an \texttt{[AND]} separator token.

The model was trained to produce all facts rated as relevant by expert raters for a given question.  This could result in long strings, so we trained two independent subtasks: a \textsc{Core} subtask that includes all facts rated \textit{important} or greater, and an \textsc{Ext} subtask that includes only \textit{extra-detail} ($0<TR<2$) facts.  The model was implemented using Huggingface Transformers \cite{wolf-etal-2020-transformers}.
% (TODO: Move to Appendix), and trained until performance plateaued at \todo{XX} epochs.  \todo{XX} permutations of fact orderings were used to discourage reliance on fact ordering and encourage robustness in model generations.  To improve inference quality, at inference time we use a batch size of 1, a beam search over 64 beams, and (given the diversity of generations) combine all facts generated in the top 10 beams (after splicing on the fact delimiter) into a candidate list.  
To simplify evaluation, T5-generated facts are aligned to their best-scoring WorldTree knowledge base fact using ROUGE-1 scores \cite{lin-hovy-2003-automatic}. %% \footnote{UQA-generated facts that do not align with WorldTree facts with a minimum ROUGE-1 score of 0.70 are discarded, as these are typically low-quality generations.} 

\paragraph{TFR-BERT (Ranking): }
A Tensorflow-Ranking-BERT model \cite{han2020learning}, trained on expert-generated data, as described in Section~\ref{sec:explanation-as-ranking}.  To move from ranking to explanation generation, we simply take the top-K ranked facts per question as the explanation, using an empirically determined threshold of $K=8$ (where F1 performance plateaued on the development set). 

\paragraph{\textsc{CoSaTa} (Schema): }
A schema-based model implemented using the Constraint Satisfaction over Tables (\textsc{CoSaTa}) solver \cite{jansen-2020-cosata}.  Schemas take the form of constraint satisfaction patterns over Worldtree facts represented as semi-structured table rows, where valid solutions of a given schema are only possible if all slots (facts) in a schema can be successfully populated by satisfying their constraints (see Figure~\ref{fig:overview}).  We use the 385 science-domain schema included with \textsc{CoSaTa}, each containing an average of 12 facts (before filtering), and run these over the WorldTree knowledge base, generating a large set of 593k solutions that are pre-cached for speed. Each solution is scored using exhaustive BERT rankings to select 1, 2, or 3 schemas to combine and output to the user.  Before output, low-scoring facts are filtered to improve succinctness.% \todo{Probably need more details to ground this one.  Also discuss scoring.}. 

% (TODO: Move to Appendix) Scoring schema to create a shortlist of relevant patterns is challenging.  Pilot experiments showed the simplest strategy had the best performance:  For a given question, we score a given schema solution by summing the BERT scores of the individual rows, while clipping any row scores below a threshold so as not to heavily penalize a given solution for a small number of irrelevant rows.  We report scores for systems that include a single schema solution, or that combine 2 or 3 top-scoring schemas.  Scores are reported for two methods: Schema$_{U}$, which provides the raw, unfiltered, and often large solutions to the user, and Schema$_{F}$, which filters any individual rows with scores below a threshold,\footnote{Here, both clipping and filtering thresholds were set at zero.} generally significantly increasing relevance while slightly decreasing completeness.\footnote{Negative results: When scoring schema solutions, pilot experiments showed a variety of different scoring methodologies based on top-K scoring, or cued scoring from other methods (e.g. T5, TFR-BERT) performed worse than a simple thresholded sum.  We hypothesize TFR-BERT ranking scores performed worse than BERT classification scores due to the comparative robustness of combining classification scores versus ranks.}

%
%   Automatic vs Manual evaluation comparison
%
\begin{table*}[t]
    \footnotesize
    \centering
    %\ra{1}
    \setlength{\tabcolsep}{3pt}
    \begin{tabular}{lccccccccccccccc}
    \toprule
    & ~& \multicolumn{5}{c}{Automatic Analysis} & ~ &  \multicolumn{3}{c}{Manual Analysis} & ~ & \multicolumn{3}{c}{Underestimate ($\Delta$)} \\
    Pattern Scoring Method & ~~~~~ & Rel & Comp & ${F1}^{ex}$ & Comp$_{B}$ &  ${F1}^{ex}_{B}$ & ~~~~ &  Rel & Comp$_{B}$ & ${F1}^{ex}_{B}$ & ~~~~~ & Rel & Comp$_{B}$ & ${F1}^{ex}_{B}$\\
    \midrule
    T5-UQA-3B$_{CORE}$                  & ~ &  0.53  &   0.36 &   0.43 & 0.10 & 0.17 &  ~ &   0.82 &  0.44  &   0.57 & ~ & \textbf{+0.29} & +0.34 & \textbf{+0.40} \\
    TFR-BERT                            & ~ &  0.72  &   \textbf{0.59} &   \textbf{0.65} & \textbf{0.36} & \textbf{0.48} &  ~ &   \textbf{0.93} &  \textbf{0.72}  &   \textbf{0.81} & ~ & +0.21 & \textbf{+0.36} & +0.33 \\
    Schema (3 Schemas)            & ~ &  \textbf{0.74}  &   0.46 &   0.57 & 0.21 & 0.33 &  ~ &   0.79 &  0.44  &   0.57 & ~ & +0.05 & +0.23 & +0.24 \\
    \bottomrule
    \end{tabular}
    \caption{\footnotesize A manual analysis of \textit{relevance} and \textit{completeness} of three length-matched single models on 50 questions from the development set.  Here, each model produces an explanation with an average length of 8 facts.  \textit{Underestimate} refers to difference scores between manual and automatic measures. \textit{Note:} Automatic analysis numbers differ from those in Table~\ref{tab:model-results} as they represent performance on only the 50 questions included in this analysis.}
    \label{tab:manual-analysis}
\end{table*}

% Original (0-6) ratings
%TR & \multicolumn{1}{c}{Fact} \\
%5 & * A light bulb generates visible light when turned on. \\
%4 & Visible light is a kind of light. \\
%6 & Light can travel through a vacuum. \\
%4 & * Light is a kind of wave. \\
%\midrule
%\multicolumn{2}{c}{Schema-Generated Explanation}\\
%\midrule
%\multicolumn{2}{l}{\textit{Schema 1: Light Properties}}\\
%6 & Electromagnetic waves can travel through a vacuum. \\
%6 & (Light is a kind of electromagnetic radiation) \\
%2 & Light travels fastest through a vacuum. \\
%4 & * Light is a kind of wave. \\[1.5mm]
%\multicolumn{2}{l}{\textit{Schema 2: Light Bulb Uses}}\\
%2 & A light bulb is used for seeing in the dark.\\
%5 & * A light bulb generates visible light when turned on.\\[1.5mm]
%\multicolumn{2}{l}{\textit{Schema 3: Parts of things}}\\
%1 & A light bulb is a part of a lamp.\\

%
%   Example also-good-but-not-automatically-identified explanation
%
\begin{table}[t]
\begin{center}
\footnotesize
\setlength{\tabcolsep}{3pt}
\begin{tabular}{cl} 
\toprule
\multicolumn{2}{c}{\textbf{Question:} Which generates waves that are capable of} \\
\multicolumn{2}{c}{traveling through a vacuum?} \\
\multicolumn{2}{c}{\textbf{Answer:} a light bulb} \\
\midrule
\multicolumn{2}{c}{Gold Explanation}\\
\midrule
TR & \multicolumn{1}{c}{Fact} \\
3 & * A light bulb generates visible light when turned on. \\
2 & Visible light is a kind of light. \\
3 & Light can travel through a vacuum. \\
2 & * Light is a kind of wave. \\
\midrule
\multicolumn{2}{c}{Schema-Generated Explanation}\\
\midrule
\multicolumn{2}{l}{\textit{Schema 1: Light Properties}}\\
3 & Electromagnetic waves can travel through a vacuum. \\
3 & (Light is a kind of electromagnetic radiation) \\
1 & Light travels fastest through a vacuum. \\
2 & * Light is a kind of wave. \\[1.5mm]
\multicolumn{2}{l}{\textit{Schema 2: Light Bulb Uses}}\\
1 & A light bulb is used for seeing in the dark.\\
3 & * A light bulb generates visible light when turned on.\\[1.5mm]
\multicolumn{2}{l}{\textit{Schema 3: Parts of things}}\\
1 & A light bulb is a part of a lamp.\\
\bottomrule
\end{tabular}
\caption{\footnotesize An example schema-generated explanation rated poorly by the automated analysis, but rated complete by the manual analysis.  \textit{TR} signifies expert (teacher) ratings of each fact.  \textit{*} signifies that a fact occurs in both generated and gold explanation.  \textit{(brackets)} signify a fact was filtered out using the schema filtering criterion to generate a more succinct explanation, but is included for demonstrative purposes.
\label{tab:example-complete-explanation}}
\end{center}
\vspace{-4mm}
\end{table}

\subsection{Automatic Evaluation Metrics}
%We use the expert relevance ratings to evaluate model \textit{relevance}, and combine these with WorldTree gold explanations to estimate \textit{completeness} automatically, or in concert with manual judgements. We also generate an F1 analogue combining the two measures. 
Here we evaluate models on \textit{relevance}, \textit{completeness}, and an F1 analogue combining the two.

%\todo{Should these be formal definitions below?}

\paragraph{Relevance:} The proportion of facts in an explanation deemed not-irrelevant to the question (i.e. that received a non-zero expert relevance rating). 

\paragraph{Completeness:} When evaluated automatically, \textit{completeness} represents the proportion of facts in the gold WorldTree explanation that are also found in the model-generated explanation.  We also define Comp$_{B}$, a binary measure, that is 1 if all facts in the WorldTree gold explanation that were rated as \textit{important} or higher by the experts are included in the model-generated explanation, and 0 otherwise. 

Due to anticipated methodological issues -- that valid explanations other than the gold annotated explanation are possible -- we also evaluate Comp$_{B}$ manually in Section~\ref{sec:manual_evaluation}, using the criterion that an experienced annotator believes the facts in the explanations form a complete chain-of-reasoning from question to answer without significant gaps.

\paragraph{F1:} With \textit{relevance} an analogue of precision, and \textit{completeness} an analogue of recall, to provide a single score that reflects overall explanation performance, we also provide an F1 analogue, defined as the harmonic mean of \textit{relevance} and \textit{completeness}:

\begin{equation}
{F1}^{ex}=\frac{Relevance \cdot Completeness}{Relevance + Completeness}
\end{equation}

\subsection{Results using Automated Metrics}
The performance of the ranking, generative, and schema-based models using the expert-informed automatic evaluation metrics is shown in Table~\ref{tab:model-results}, broken down by single models and ensembles.  In terms of single models, at 0.82 the Schema-based models perform highest in \textit{relevance} -- likely owing to the constraints of each schema ensuring that collections of facts are organized according to a particular theme -- followed by TFR-BERT, with the best performing T5-UQA model performing 20 points lower, at 0.62.  Conversely, TFR-BERT scores highly in both graded and binary \textit{completeness}, reaching 0.50 Comp$_{B}$, while T5-UQA reaches less than half this performance, and the schema models reach a middle-ground of 0.36.  The best-scoring model, TFR-BERT, reaches 0.66 ${F1}^{ex}$, or 0.37 ${F1}^{ex}_{B}$ using binary completeness. 

Given the variety of methodologies used, the three model families have comparatively low overlap, and ensemble models that combine the output of each model substantially improve \textit{completeness} and ${F1}^{ex}$ scores.  The best-scoring model, which combines the output of the generative, ranking, and schema models achieves an ${F1}^{ex}_{B}$ of 0.57, reaching 20 points over the best-scoring single model.

\subsection{Manual Evaluation of Completeness}
\label{sec:manual_evaluation}

The approximate automatic measure of \textit{completeness} used above is still problematic, because it relies on a single gold explanation filtered to include only the most important facts in the expert ratings.  To measure the difference between this automatic measure of completeness and actual completeness, we conducted a detailed manual evaluation of single model performance for 50 questions in the development set.  To control for explanation length, we chose single models whose average explanation lengths were identical ($8\pm0.5$ facts long), while for robustness we evaluated completeness using binary judgements.  In addition, any facts without expert relevance ratings (i.e. facts that were not within the initial top-K list rated by the expert annotators) were provided binary relevance judgements.\footnote{A total of 327 facts, approximately 2 per model-generated explanation, required these relevance judgements.}

\paragraph{Raters:}  Annotating the completeness of a collection of facts as an explanation can be challenging, particularly when locating gaps in an inference.  Due to timing constraints, in this analysis, completeness judgements were initially made by an author (science-domain expert and compositional reasoning expert), then compared against those generated by one of the domain-expert science teachers from Section~\ref{sec:dataset}.  Percent agreement between the author and teacher was strong, at 89\% for binary completeness judgements, and 88\% for binary relevance judgements.

\paragraph{Results:} The results of this manual analysis are shown in Table~\ref{tab:manual-analysis}.  This analysis shows that even when supplemented with expert relevance ratings, using a single gold explanation for automatically evaluating completeness still provides a large underestimate of task performance.  In particular, manual binary completeness ratings Comp$_{B}$ exceeded automatic evaluations by +23\% to +36\%, \textit{and in all cases more than doubled the original estimate of task performance}.  To illustrate this, Table~\ref{tab:example-complete-explanation} shows an example of a gold explanation, and an explanation generated by the Schema model.  Both explanations are complete, but the Schema-generated explanation is rated poorly because it includes only half of the highly-rated facts of the gold explanation.  Clearly as explanations become large, and composed of increasingly atomic facts, many more paths to generating complete explanations are possible, and alternate methods of accurately estimating completeness are required. 

%- Results on TEST from top models, and
%- Manual analysis of the same ~50(?) questions from those models. 
%- Overlap analysis?  What proportion of questions does model X do well on, that model Y does poorly on?
%\todo{ \lipsum[1-3] }

\section{Conclusion}

\paragraph{Relevance performance is still undercounted:} While the expert-generated relevance ratings produced in this work provide more accurate estimates of performance compared to single gold explanations when used in fully-automatic evaluations, these automatic estimates still undercount overall model performance.  In our experiments we show the expert ratings primarily provide a vehicle for training better models, but that automatically evaluating relevance performance still remains a challenge, even with a large targeted increase in relevance annotation. Further, annotators reported that determining  relevance of single facts in isolation is challenging because it lacks the broader compositional context of the rest of the candidate explanation, suggesting ultimate limits to the utility of exhaustive annotation. 

\paragraph{Measuring completeness is a major challenge:} As explanations become larger, and facts become more atomic, models are afforded more opportunities to build explanations that differ from those annotated as gold.  Because of this, automatic completeness judgements substantially undercount true (manual) completeness by at least a factor of two across all models.  Alignment approaches \cite{inoue-etal-2020-r4c} and annotating multiple explanations \cite{jhamtani-clark-2020-learning} have been proposed for short explanations, but are unlikely to scale well as compositionality increases.  Treatments similar to those that use a formal semantics or theorem proving to evaluate truth \citep[e.g.][]{weber-etal-2019-nlprolog,clark2020transformers} are attractive, but are unlikely to offer generality at scale without substantial development effort.  Conversely, streamlined manual evaluation frameworks are becoming increasingly common for simpler generative tasks \cite[e.g.][]{khashabi2021genie}, but it is unclear how accurately non-technical crowdworkers would perform on rating compositional completeness -- and even if possible, this would raise the time and cost associated with evaluation dramatically.

\paragraph{Automatic model comparisons are inaccurate:} While T5-UQA and the Schema models show large performance differences using automatic measures, our manual analysis shows they actually have similar performance characteristics as performance underestimates disproportionately affect different models.  Comparing models to perform hypothesis testing (i.e. \textit{Model A outperforms Model B}) is currently challenging without substantial manual analysis, and a significant methodological limitation to advancing the science of compositional reasoning for building large explanations.

\subsection*{Open Data}
Our corpus and analyses are available at: \url{http://cognitiveai.org/explanationbank/} .
%\todo{ \lipsum[1] }

\section*{Acknowledgements}
We thank the three anonymous reviewers for their helpful comments. This work supported in part by National Science Foundation (NSF) award \#1815948 to PJ. 

\bibliography{anthology,custom}
\bibliographystyle{acl_natbib}

%\begin{comment}

\appendix

\patchcmd{\quote}{\rightmargin}{\leftmargin 2mm \rightmargin}{}{}

\section{Appendix}
\label{sec:appendix}

\subsection{Data}

\subsubsection{WorldTree V2 explanation corpus}

The WorldTree V2 explanation corpus \cite{xie-etal-2020-worldtree} is a set of explanations to standardized science exam questions represented as sets of atomic facts.  Each fact in an explanation is connected with either the question, answer, or other facts using lexical overlap (shared words).  The average explanation contains 6 facts (range 1-16).  The supporting semi-structured knowledge base contains approximately 9k facts distributed across 82 tables, where each table is organized around a particular kind of knowledge relation (e.g. \textit{taxonomic}, \textit{parts-of}, \textit{generic properties},  \textit{sources of things}, \textit{causes}, \textit{changes}, \textit{if-then relationships}, \textit{coupled relationships}, etc). 

WorldTree V2 includes a subset of the questions from the Aristo Reasoning Challenge (ARC) corpus \cite{clark2018think}.  ARC questions are standardized science exam questions drawn from 12 US states, where each question is a 4-choice multiple choice question. In total, WorldTree V2 contains 2210 training, 496 development, and 1670 test set questions. 

\subsection{Expert Relevance Ratings}

A total of 236k expert judgements were collected for 126k facts.  While most facts were rated by 2 annotators, due to scheduling conflicts with the COVID-19 pandemic, approximately 15\% of questions are rated by a single teacher.  120 randomly sampled questions (approximately 3.5k facts) were used as training and calibration for all 3 annotators, who iteratively rated facts then  discussed and resolved disagreements as a means of calibration before annotating the remainder of questions. 

\paragraph{Exclusions:} The WorldTree V2 knowledge base contains approximately 1.2k synonymy relations (e.g. \textit{cooler} means \textit{colder}, or \textit{bike} means \textit{bicycle}) that models tend to rate highly even though they have minimal conceptual content.  We filtered these synonymy facts from the facts that the expert annotators rated before assembling the shortlists, to ensure that expert time was spent on rating the relevance of core scientific/world knowledge rather than thesaurus-like facts.

%\subsection{Additional Binary Relevance Ratings}
%Additional binary relevance ratings were collected in Section~\ref{sec:manual_evaluation} for facts that were chosen by a given model, but were not rated by the domain experts (i.e. the fact didn't fall within the top-K shortlist for expert rating described in Section~\ref{sec:dataset}).  After filtering, a total of 327 facts (an average of 6.5 facts per question across all three models) required additional manual relevancy ratings. 

\subsection{Evaluation Metrics}

\subsubsection{Ranking Metrics}

Mean Average Precision (MAP) and Normalized Discounted Cumulative Gain (NDCG) follow their standard definitions as used in Table~\ref{tab:ranking-results}.

\subsubsection{Whole-explanation Metrics}

\paragraph{Relevance:} Relevance represents the proportion of facts returned by a model that have a non-zero expert-rated relevance score for a given question.  More specifically, relevance for a given explanation of length $L$ is defined as:

\begin{equation}
    Relevance = \frac{\sum_{i=1}^{L} {R}_{i}}{L}
\end{equation}

Where $TR(i, q)$ is an expert-annotated relevance rating for fact $i$ for question $q$, the relevance score of a given fact in the explanation, ${R}_{i}$, is defined as:

\begin{equation}
    %\[
    {R}_{i} = 
\begin{cases}
    1,              & \text{if } TR(i, q) \geq 1\\
    0,              & \text{otherwise}
\end{cases}
%\]
\end{equation}
Note that facts without annotated relevance ratings are assumed to have a rating of 0 (irrelevant).

\paragraph{Completeness:} Completeness represents the proportion of facts in the gold explanation for a given question that are also present in the model-generated explanation.  Given a set of facts representing a gold explanation of length $N$, $G = \{G_1, G_2, ..., G_N\}$, and set of facts representing the model-generated explanation $M = \{M_1, M_2, ..., M_L\}$, the completeness of $M$ is defined as:

\begin{equation}
    Completeness = \frac{|G\cap M|}{|M|}
\end{equation}

The binary measure of completeness, Comp$_B$, is 1 if \textit{Completeness} is 1, and 0 otherwise.

\subsection{Additional Model Details and Hyperparameters}

\subsubsection{T5-UnifiedQA} 
UnifiedQA \cite{khashabi-etal-2020-unifiedqa} is a variant of T5 \cite{2020t5} that includes multi-task pretraining for 4 different forms of question answering across 20 datasets, including extractive QA (i.e., locating an answer span in a passage), abstractive QA (i.e., generating an answer not supplied in a passage), multiple-choice QA, and Boolean QA, while achieving state-of-the-art performance on 10 datasets.  Here we train T5-UQA-3B to also generate compositional explanations by cueing generation with the question and correct answer candidate, and targeting generation to produce long strings of highly-rated facts delimited with an \texttt{[AND]} separator token.

\paragraph{Cueing:} For training and evaluation, data was provided in the following format.  Source (question) data was provided in the following format: 
\begin{quote}
\texttt{explanation: <question text> [ANSWER] <answer text> [CUETOKEN]}
\end{quote}
Target (explanation) data was provided and generated in the following format:
\begin{quote}
\texttt{<fact1> [AND] <fact2> [AND] <fact3> [AND] ... [EOS]}
\end{quote}
Where \texttt{factN} represents the sentence tokens for a given fact in the explanation (e.g. \textit{``water is a kind of liquid''}). 5 permutations of fact orderings were used to discourage reliance on fact ordering and encourage robustness in model generations.

\paragraph{Pretraining:} During a pretraining phase, T5-UQA was cued with the question and answer (as above), but provided with only a single fact from the gold explanation to generate. 

\paragraph{Model parameters:} All experiments were performed with the 3-billion parameter version of T5-UQA.  We made use of DeepSpeed ZeRo optimizations \cite{rajbhandari2020zero} to fit the 3B model into the largest GPUs available to us (A100-40GB).  Models were trained to 30 epochs, where generation performance (ROUGE-1) plateaued. We use the default hyperparameters for training provided in the Huggingface Transformers library \cite{wolf-etal-2020-transformers}.  To improve inference quality, at inference time we use a batch size of 1, a beam search over 64 beams, and (given the diversity of generations, and the preference for shorter generations even after considerable training) combine all facts generated in the top 10 beams (after splicing on the fact delimiter) into a candidate list of generated facts.

\paragraph{Model runtime:} T5-UQA-3B took approximately 3 days to train on our dataset and another 2 days to evaluate, using 4x A100-40GB GPUs (i.e. approximately 20 A100 GPU days, equivalent to approximately 52 V100 GPU days). 

\paragraph{Alignment to WorldTree knowledge base:} To enable automatic evaluation and direct comparison with the other models, the output of T5-UQA was aligned to existing WorldTree facts.  The output of the model was split on the \texttt{[AND]} delimiter, and each fact was exhaustively scored against all 9k facts in the WorldTree tablestore, where the fact with the highest ROUGE-1 \cite{lin-hovy-2003-automatic} alignment score was taken to be the appropriate WorldTree fact.  We empirically determined that facts whose ROUGE-1 scores were lower than 0.70 tended to be poor alignments, most typically from misgenerations, incorrect generations, or nonsensical generations, though occasionally from correct generations that do not have a corresponding counterpart in the WorldTree knowledge base. 

\subsubsection{TFR-BERT}
A Tensorflow-Ranking-BERT model \cite{han2020learning}, which combines large language model embeddings with pointwise ranking (rather than classification) through the Tensorflow-Ranking framework \cite{TensorflowRankingKDD2019}.

\paragraph{Cueing:} During training and evaluation, the model was provided both the question and correct answer text.  During training, it was provided with relevance rankings for the shortlist (approximately top-30) expert-rated facts.  During evaluation, it was provided the top-100 ranked facts from the exhaustive BERT baseline (ranked by their classification scores), and re-ranked these facts. 

\paragraph{Model parameters:} Due to a large memory dependency (TFR-BERT GPU RAM scales with model parameter size \textit{and} list size), we made use of BERT-base-uncased, a 110M parameter model.  Default parameters were used for training and evaluation.  

\paragraph{Model runtime:} Training took approximately 2 days on a single A100-40GB GPU (multi-GPU training is not currently supported).  To evaluate on comparatively large list sizes, evaluation was done using CPUs rather than GPUs, and took approximately 3 days using 32 CPU cores. 

\paragraph{Top-K tuning:} TFR-BERT results are reported both as a ranking model, where the output is a ranking of the entire knowledge base, as well in a whole-explanation paradigm where the output is a discrete \textit{top-k-fact} explanation.  Here we choose $k=8$, where ${F1}^{ex}$ performance plateaued at 0.65 on the development set. 

\subsubsection{CoSaTa Schemas}
A schema-based model implemented using the Constraint Satisfaction over Tables (\textsc{CoSaTa}) solver \cite{jansen-2020-cosata}.  Schemas take the form of constraint satisfaction patterns over Worldtree facts represented as semi-structured table rows, where valid solutions of a given schema are only possible if all slots (facts) in a schema can be successfully populated by satisfying their constraints.  We use the 385 science-domain schema included with \textsc{CoSaTa}, each containing an average of 12 facts, and run these over the WorldTree tablestore, generating a large set of 593k solutions that are pre-cached for speed. 

\paragraph{Scoring:} Scoring schema to create a shortlist of relevant patterns is challenging.  Pilot experiments showed the simplest strategy had the best performance:  For a given question, we score a given schema solution by summing the BERT scores of the individual facts used in that solution, while clipping any fact scores below a threshold so as not to heavily penalize a given solution for a small number of irrelevant rows.  We report scores for systems that include a single schema solution, or that combine 2 or 3 top-scoring schemas.  
%Scores are reported for two methods: Schema$_{U}$, which provides the raw, unfiltered, and often large solutions to the user, 
Reported performance is for a model that post-filters schemas before giving them to the user. Specifically, any facts in a schema with scores below a threshold are filtered,\footnote{Here, both clipping and filtering thresholds were set at zero.} generally significantly increasing relevance while slightly decreasing completeness. \textit{Negative results:} When scoring schema solutions, pilot experiments showed a variety of different scoring methodologies based on top-K scoring, or cued scoring from other methods (e.g. T5, TFR-BERT) performed worse than a simple thresholded sum.  We hypothesize TFR-BERT ranking scores performed worse than BERT classification scores due to a comparative robustness of combining multiple classification scores into a single score, versus combining multiple ranks for individual facts.

\paragraph{Model parameters:} Scoring for CoSaTa schemas uses the BERT-base-uncased (110M) model. Other hyperparameters reported above.

\paragraph{Model runtime:} Initial schema generation and caching took approximately 2 days on 16 CPU cores.  Subsequent ranking, scoring, and filtering took approximately 2 hours. 

\subsubsection{BERT and RoBERTa Baselines}

The BERT and RoBERTa exhaustive baselines are trained as classifiers, using a distant supervision paradigm where (for a given question) all gold explanation facts are taken as gold, and $N$ randomly sampled facts from the knowledge base are taken as negative examples.  Unlike Das et al. \cite{das2019chains}, who train an exhaustive model using all non-gold facts in the knowledge base, here we subsample only a subset of the corpus to minimize the likelihood that the model would use a relevant fact not annotated as in the gold explanation for a given question as a negative example. 

\paragraph{Model Parameters:} We used a vanilla BERT-base-uncased (110M), as well as RoBERTA-Large (355M) pre-trained on the RACE reading comprehension benchmark \cite{lai2017race}.

%
%This is an appendix.

%\end{comment}

\end{document}